\documentclass{article}
\usepackage{amsmath,amssymb,graphicx,comment,subcaption}
\usepackage{color,url}
\usepackage{authblk}
\usepackage[ruled,norelsize]{algorithm2e}

\newcommand{\note}[1]{{\textbf{\color{red}#1}}}
\title{A Quasi-Newton algorithm on the orthogonal manifold for NMF with transform learning}
\author[$\dagger$]{Pierre Ablin}
\author[$\ddagger$]{Dylan Fagot}
\author[$\ddagger$]{Herwig Wendt}
\author[$\dagger$]{Alexandre Gramfort}
\author[$\ddagger$]{C\'edric F\'evotte}
\affil[$\dagger$]{Inria, Parietal team, Universit\'e Paris-Saclay, Saclay, France}
\affil[$\ddagger$]{IRIT, Universit\'e de Toulouse, CNRS, Toulouse, France}
\newcommand{\ve}[1]{ {\mathbf{#1}} }
\usepackage{soul,ulem}
\begin{document}
\maketitle
\thanks{$^\dagger$ Supported by the Center for Data Science, funded by the IDEX Paris-Saclay, ANR-11-IDEX-0003-02, and the European Research Council (ERC SLAB-StG-676943). $^\ddagger$ Supported by the European Research Council (ERC FACTORY-CoG-6681839).}
\begin{abstract}
Nonnegative matrix factorization (NMF) is a popular method for audio spectral unmixing.
While NMF is traditionally applied to off-the-shelf time-frequency representations based on the short-time Fourier or Cosine transforms, the ability
to learn transforms from raw data attracts increasing attention.
However, this adds an important computational overhead.
When assumed orthogonal (like the Fourier or Cosine transforms), learning the transform yields a non-convex optimization problem on the orthogonal matrix manifold.
In this paper, we derive a quasi-Newton method on the manifold using sparse approximations of the Hessian.
Experiments on synthetic and real audio data show that the proposed algorithm outperforms state-of-the-art first-order and coordinate-descent methods by orders of magnitude.
A Python package for fast TL-NMF is released online at {\footnotesize \tt \url{https://github.com/pierreablin/tlnmf}}.
\end{abstract}
\section{Introduction}
\label{sec:intro}
Nonnegative matrix factorization (NMF) consists in decomposing a nonnegative data matrix $\mathbf{V}\in\mathbb{R}_+^{M\times N}$ into
\begin{equation}
\mathbf{V} \approx \mathbf{W}\mathbf{H}
\label{eq:nmf}
\end{equation}
where $\mathbf{W}\in\mathbb{R}_+^{M\times K}$ and $\mathbf{H}\in\mathbb{R}_+^{K\times N}$ are two nonnegative matrices referred to as \emph{dictionary} and \emph{activation} matrix, respectively.
The rank $K$ of the factorization is generally chosen to be smaller than $\text{min}(M,N)$ so that the approximation is low-rank.
This method has been popularized by the seminal paper of Lee and Seung~\cite{lee1999learning}.
In audio signal processing, $\mathbf{V}$ is typically a magnitude $|\ve{X}|$ or power $|\mathbf{X}|^{\circ 2}$ spectrogram, where $\ve{X}$ is the short-time Fourier or Cosine transform of some signal $y(t)$ (the notation $\circ$ denotes element-wise operations throughout the paper).
%
%
The short-time frequency transform $\ve{X}$ is computed by applying an orthogonal frequency transform $\mathbf{\Phi}$ to the \emph{frames matrix} $\mathbf{Y}\in\mathbb{R}^{M\times N}$ which contains windowed segments of the original temporal signal $y(t)$ in its columns.
%
%
$M$ is the length of the window and $N$ is the resulting number of time frames.
As such, we have $\mathbf{X}=\mathbf{\Phi}\mathbf{Y}$.
Factorizing $\mathbf{V}$ as in (\ref{eq:nmf}) can lead to a meaningful decomposition where the dictionary $\mathbf{W}$ captures spectral patterns and the activation matrix $\mathbf{H}$ contains data decomposition coefficients.
This decomposition can then be used to solve a variety of signal processing problems such as source separation \cite{smaragdis2014static,vincent2018audio,cichocki2009nonnegative} or music transcription \cite{smaragdis2003non,vincent2008harmonic}.
\note{}
In the latter works, $\ve{V}$ is computed with a given off-the-shelf short-time frequency transform.
This sets a limit to the accuracy of the factorization.
To adress this issue, transform-learning NMF (TL-NMF) was introduced in \cite{fagot2018transform,wendt2018jacobi}. It computes an optimal transform from the input signal: the transform $\mathbf{\Phi}$ is learned \emph{together} with the latent factors $\mathbf{W}$ and $\mathbf{H}$.
TL-NMF has been employed successfully for source separation examples: it leads to better or comparable performance as compared with traditional fixed-transform NMF \cite{fagot2018transform,wendt2018jacobi,Yoshii2018a}.

The contribution of this article is to propose a faster solver for TL-NMF. In~\cite{fagot2018transform}, an orthogonal transform is learned using a projected gradient descent onto the orthogonal matrix manifold.
In~\cite{wendt2018jacobi}, a faster Jacobi approach (in which $\ve{\Phi}$ is searched as a product of Givens rotations) is proposed.
In a different framework, \cite{Yoshii2018a} optimizes a nonsingular transform (not constrained to be orthogonal) with majorization-minimization (MM).
%
In all cases, the cost of TL-NMF remains prohibitively large compared to standard NMF.
The estimation of the transform is the computational bottleneck of the algorithms, and takes orders of magnitude more time than standard NMF.
The present work aims at reducing the gap in terms of execution time between TL-NMF and traditional NMF in the orthogonal transform setting (which gently relaxes Fourier or Cosine transforms while still imposing orthogonality). To that purpose, we propose a quasi-Newton method on the orthogonal manifold.

The article is organized as follows.
Section~\ref{sec:tl} introduces the optimization problem behind TL-NMF and presents the standard MM updates used for $\ve{W}$ and $\ve{H}$.
Section~\ref{sec:optim} starts with a brief introduction to optimization on the orthogonal manifold, introduces previous work and presents the new quasi-Newton algorithm.
Finally, Section~\ref{sec:expe} describes comparative experiments with synthetic and real data.
%
Exploiting the reduced computational load, we highlight a previously unnoticed energy concentration phenomenon of the learned transform, and study the structure of the local minima of the objective function.

\noindent{\bf Notation.\quad} Scalars are written in lower-case (e.g., $v\in\mathbb{R}$), vectors in bold lower-case (e.g., $\mathbf{v}\in\mathbb{R}^M$) and matrices in bold upper-case (e.g., $\mathbf{V}\in\mathbb{R}^{M\times N}$), while tensors are in calligraphic upper-case (e.g., $\mathcal{H}\in\mathbb{R}^{M\times M \times M \times M}$).
Entry $(m,n)$ of a matrix $\mathbf{V}$ is denoted as $v_{mn}$ or $[\mathbf{V}]_{mn}$ while entry $(i,j,k,l)$ of a tensor $\mathcal{H}$ is denoted as $\mathcal{H}_{ijkl}$.
The identity matrix of size $M$ is denoted as $\mathbf{I}_M $.
The element-wise operations between two matrices $\mathbf{A}$ and $\mathbf{B}$ are written $\mathbf{A}\circ\mathbf{B}$ and $\frac{\mathbf{A}}{\mathbf{B}}$ for the multiplication and division while $\mathbf{A}^{\circ p}$ and $|\mathbf{A}|$ denote the element-wise exponentiation and modulus, respectively.
The orthogonal matrix set $\mathcal{O}_M$ is the set of matrices such that ${\bf M M^{\top}= I}_M$.
The Frobenius scalar product is denoted as $\langle  \mathbf{A} \lvert \mathbf{B} \rangle = \sum_{i, j} a_{ij}b_{ij}$.
Given a fourth-order tensor $\mathcal{H}$ of size $M \times M \times M \times M$, the weighted Frobenius inner product is $ \langle \mathbf{A} \lvert \mathcal{H} \rvert \mathbf{B} \rangle = \sum_{i, j, k, l}\mathcal{H}_{ijkl} a_{ij}b_{kl}$.
The Itakura-Saito divergence is given by $d_{\text{IS}}(x, y) = \frac{x}{y} - \log(\frac{x}{y}) - 1$.
Finally, $\delta_{ij}$ is the Kronecker delta function of $(i,j)$ equal to $1$ if $i=j$ and $0$ otherwise.

\section{NMF with transform learning}
\label{sec:tl}

\subsection{Objective function}
\label{seq:tl_nmf}
TL-NMF consists in solving a NMF problem while learning a data-adapted transform \cite{fagot2018transform}. This is done by minimizing some measure of fit between the transformed data
$|\mathbf{\Phi Y}|^{\circ 2}$ and the factorized expression $\mathbf{WH}$ where we here assume that $\ve{\Phi}$ is a real-valued orthogonal matrix (of size $M \times M$). In addition, a penalty is added to promote sparsity of the activation coefficients. The TL-NMF problem thus writes:
\begin{align}
\label{eqn:TLobj}
&\min_{\mathbf{\Phi}, \mathbf{W},\mathbf{H}} \mathcal{C}_{\lambda}({\bf\Phi, W, H}) =  D_{IS}( | \mathbf{\Phi} \mathbf{Y} |^{\circ 2} | \mathbf{W} \mathbf{H} )+\lambda \frac{M}{K}||\mathbf{H}||_1 \nonumber\\
&\text{s.t.}\enspace  \mathbf{W} \ge 0, \mathbf{H} \ge 0, \ \forall k,   ||\mathbf{w}_k||_1=1, \mathbf{\Phi} \mathbf{\Phi}^{\top} = \mathbf{I}_{M},
\end{align}
where $\mathbf{w}_k$ is the $k$-th column of $\bf W$ and $D_{IS}(\cdot|\cdot)$ is the Itakura-Saito (IS) divergence defined as $D_{IS}(\mathbf{A}|\mathbf{B}) = \sum_{m,n}  d_{IS}(a_{mn}|b_{mn}) = \sum_{m,n} \frac{a_{mn}}{b_{mn}} - \log\frac{a_{mn}}{b_{mn}} -1 $. Note that any other measure of fit could be used with no loss of generality. However, the IS divergence has been proven to be particularly relevant for decomposing power spectrograms~\cite{fevotte2009nonnegative}. The $M/K$ factor allows to the values of the measure of fit and of the penalty term to be of comparable orders of magnitude.

\begin{algorithm}[tb]
\SetKwInOut{Input}{Input}
\SetKwInOut{Output}{Output}
 \Input{Frames matrix $\mathbf{Y}$, dictionary size $K$, minimization algorithm for transform learning $\mathcal{A}$, number of iterations of the TL minimization $L$, total number of iterations $N_{\text{it}}$}
 Initialize $\bf\Phi, W, H$. \\
 \For{$n=1,\cdots, N_{\text{it}}$}{
    \underline{\textbf{NMF}} \\
	Compute the current spectrogram $\mathbf{V} = \lvert \mathbf{\Phi Y} \lvert ^{\circ 2}$ \\
	Decrease $\mathcal{C}_{\lambda}$ w.r.t. $\bf (W, H)$ (step $\star$)\\
    \underline{\textbf{TL}} \\
	Compute $\bf \hat{V}=WH$ \\
	Update $\Phi \leftarrow \mathcal{A}({\bf\hat{V}, Y,  \Phi},L)$\\
  }
 \Output{ $\bf\Phi, W, H$}
 \caption{Alternate minimization for TL-NMF}
 \label{algo:alternate}
\end{algorithm}

A straightforward estimation procedure to solve the problem~\eqref{eqn:TLobj} is to use \emph{alternate minimization}.
It is summarized in Algorithm~\ref{algo:alternate}.
It alternates between two steps.
In the NMF step, the current ``spectrogram'' $\mathbf{V} = |{\bf \Phi Y}|^{\circ 2}$ is fixed and the algorithm decreases $\mathcal{C}_{\lambda}$ with respect to (w.r.t.) $\mathbf{W}$ and $\mathbf{H}$.
This is done using classical NMF MM update rules, described in the next section.
In the transform-learning part, the factorization $\hat{\mathbf{V}} = \bf WH$ is fixed, and the algorithm decreases $\mathcal{C}_{\lambda}$ w.r.t. $\mathbf{\Phi}$, using an optimization algorithm denoted as $\mathcal{A}$.
This article aims to propose a new fast algorithm $\mathcal{A}$ for the minimization of $\mathcal{C}_{\lambda}$ w.r.t. $\mathbf{\Phi}$.

\subsection{Majorization-minimization updates of $\ve{W}$ and $\ve{H}$}
\label{sec:nmfrules}

We update $\mathbf{W}$ and $\mathbf{H}$ (step $\star$ in  Algorithm~\ref{algo:alternate}) with the standard multiplicative updates that can be derived from a majorization-minimization procedure \cite{fevotte2011algorithms}. The sum-to-one constraint on the columns of $\mathbf{W}$ (which is necessary to avoid degenerate solutions) can be rigorously enforced using a change of variable, like in \cite{lefevre2011itakura,essid2013smooth}. The updates read:
$$
\mathbf{H}\gets\mathbf{H}\circ\left[\frac{\mathbf{W}^T\left((\mathbf{WH})^{\circ -2}\circ |\mathbf{\Phi Y}|^{\circ 2}\right) }{\mathbf{W}^T(\mathbf{WH})^{\circ -1}+ \lambda\frac{M}{K}\mathbf{1}_{K \times N}}\right]^{\circ\frac{1}{2}},
$$
%
$$
\mathbf{W}\gets\mathbf{W}\circ\left[\frac{\left((\mathbf{WH})^{\circ -2}\circ |\mathbf{\Phi Y}|^{\circ 2}\right) \mathbf{H}^T}{(\mathbf{WH})^{\circ -1}\mathbf{H}^T + \lambda\frac{M}{K} \mathbf{1}_{M \times N}\mathbf{H}^T}\right]^{\circ\frac{1}{2}}.
$$
They should be followed by a joint normalization of the columns of $\ve{W}$ and rows of $\ve{H}$ \cite{lefevre2011itakura,essid2013smooth}.

\section{Quasi-Newton update of the transform $\mathbf{\Phi}$}
\label{sec:optim}
\subsection{Optimization on the orthogonal manifold}
\label{sec:optim_on}

This section focuses on the minimization of $\mathcal{C}_{\lambda}$ with respect to $\mathbf{\Phi}$.
In the following, we define $\bf \hat{V} = WH$, and let $\mathcal{L}({\mathbf{\Phi}}) = D_{IS}({\bf\lvert \Phi Y \rvert}^{\circ 2} \lvert \bf \hat{V}) $.
We may write:
%
\begin{equation}
\mathcal{L}(\mathbf{\Phi}) = \sum_{m=1}^M\sum_{n=1}^N f_{\hat{v}_{mn}}([\mathbf{\Phi} \mathbf{Y}]_{mn}),
\label{eq:cost}
\end{equation}
where we define $f_v(x) = d_{IS}(x^2, v)= \frac{x^2}{v} - 2\log(\frac{x}{y}) - 1$.
The orthogonality constraint imposed to $\mathbf{\Phi}$ ($\mathbf{\Phi} \mathbf{\Phi}^{\top}=\mathbf{I}_M$)
implies that \eqref{eq:cost} should be minimized on the \emph{orthogonal matrix manifold} $\mathcal{O}_M$.
This manifold appears in many optimization problems and its geometry is well-studied~\cite{absil2009optimization}.
To derive an iterative algorithm that minimizes \eqref{eq:cost}, we propose to parametrize the neighborhood of an iterate $\mathbf{\Phi}^t$ via the \emph{matrix exponential} (following, e.g., \cite{ablin2018orthogonal}).
We set:
\begin{equation}
    \mathbf{\Phi}^{t+1} = \exp(\mathbf{E})\mathbf{\Phi}^t,
\label{eq:param}
\end{equation}
where $\mathbf{E}$ is an anti-symmetric matrix.
If $\mathbf{\Phi}^t $ is orthogonal, this update enforces that $\mathbf{\Phi}^{t+1}$ remains orthogonal.
It thus provides a natural framework for iterative optimization over the orthogonal manifold.

\subsection{Previous methods}
\label{sec:previous}
A projected gradient method is presented in~\cite{fagot2018transform}.
Iterates are of the form:
\begin{equation}
 \mathbf{\Phi} \leftarrow \Pi((\mathbf{I}_M - \eta \mathbf{G}) \mathbf{\Phi} ),
\label{eq:proj_grad}
\end{equation}
where $\mathbf{G}$ is the \emph{natural gradient}~\cite{amari1998natural} of $\mathcal{L}$, $\eta$ is a step-size, and $\Pi$ is the projection to the manifold, given by $\Pi(\mathbf{C}) = (\mathbf{C}\mathbf{C}^{\top})^{-\frac12}\mathbf{C}$.
The main drawback is that, as a first order method, it is hard to have a proper step size policy, and the convergence is at most linear~\cite{nocedal1999optim}.

A variant was proposed in \cite{wendt2018jacobi} where the transform was updated using Givens rotations as:
\begin{equation}
    \mathbf{\Phi} \gets \mathbf{R}_{pq}(\theta)\mathbf{\Phi}
\end{equation}
where $\mathbf{R}_{pq}$ is a unidirectional rotation matrix with axis $(p,q)$ and angle $\theta$.
This update rule results in an acceleration because the single-axis rotations are cheap to compute.
However, finding the best angle $\theta$ given an axis $(p,q)$ was shown to involve a highly non-convex problem with the presence of many local minima. As such $\theta$ is selected by grid search which is not entirely satisfactory.
%
%


\subsection{Derivatives of the objective function}
\label{sec:der}
In this section, the derivatives of $\mathcal{L}$ with respect to the parametrization~\eqref{eq:param} are computed.
The gradient is a $M \times M$ matrix denoted as $\mathbf{G}$, and the Hessian is a $M \times M \times M \times M$ tensor denoted as $\mathcal{H}$.
They are obtained from the following second-order Taylor expansion:
\begin{equation}
\mathcal{L}(\exp(\mathbf{\mathbf{E}})\mathbf{\Phi}) = \mathcal{L}(\mathbf{\Phi}) + \langle \mathbf{G} \lvert \mathbf{E}\rangle + \frac{1}{2} \langle \mathbf{E} \lvert \mathcal{H} \rvert  \mathbf{E} \rangle +  \mathcal{O}(\lvert \lvert \mathbf{E} \rvert \rvert ^3).
\label{eq:taylor}
\end{equation}
Using $\mathbf{X} = \mathbf{\Phi} \mathbf{Y}$, the gradient is given by
%
\begin{equation}
\mathbf{G}_{ij} = \sum_{n=1}^N f'_{\hat{v}_{in}}(x_{in})x_{jn} = 2\sum_{n=1}^N (\frac{x_{in}}{\hat{v}_{in}} -\frac{1}{x_{it}})x_{jt}
\label{eq:gradient}
\end{equation}
and the Hessian is given by
%
\begin{equation}
\mathcal{H}_{ijkl} = \delta_{ik} \sum_{n=1}^N f''_{\hat{v}_{in}}(x_{in})x_{jn}x_{ln} + \delta_{jk} \mathbf{G}_{il}.
\label{eq:hessian}
\end{equation}

\noindent\textbf{Newton's method.\quad} Newton method on the manifold would take $\mathbf{E} = -\Pi_A(\mathcal{H}^{-1} \mathbf{G})$, where $\Pi_A$ is the projection onto the anti-symmetric matrices:
%
\begin{equation}
\Pi_A(\mathbf{C}) = \frac{\mathbf{C} - \mathbf{C}^{\top}}{2}.
\label{eq:proj}
\end{equation}
Note that this projection is much cheaper to compute than $\Pi$.
Newton's method provides fast convergence, but is not practical for several reasons.
First, it requires the computation of the Hessian.
The complexity of this operation is $O(M^3 \times N)$. Besides, the cost of computing a gradient is $O(M^2 \times N)$.
Thus, a gradient method can roughly perform $M$ iterations when Newton's method performs one.
Second, because the problem is non-convex, the Hessian should be regularized to enforce its positive-definiteness, thereby guaranteeing that $-\mathcal{H}^{-1}\mathbf{G}$ is a descent direction.
A standard regularization procedure
consists in adding $\mu \mathbf{I}$
to the Hessian where
$\mu > \max(0,-\lambda_{\text{min}})$ and where $\lambda_{\text{min}}$ is the smallest eigenvalue of $\mathcal{H}$.
The Hessian is sparse, but its sparsity structure does not help in computing the key quantity $\lambda_{\text{min}}$.
As such one we would have to compute the smallest eigenvalue of a $M^2 \times M^2$ matrix which is prohibitively expensive.
Finally, solving the $M^2 \times M^2$ linear system $\mathcal{H}\mathbf{E} = -\mathbf{G}$ using, e.g., Gaussian elimination has complexity $O(M^6)$, which is orders of magnitude higher than the computation of the gradient.

\subsection{A fast algorithm based on Hessian approximation}
\label{sec:algo}

To derive a practical \emph{quasi-Newton} algorithm, one can observe that
the Hessian of $\mathcal{L}$ has two terms.
The second term, $ \delta_{jk} \mathbf{G}_{il}$, cancels when the algorithm is close to convergence, so we may ignore it.
As an approximation of the first term, we impose that it cancels when $j \neq l$, leading to the following Hessian approximation:
%
\begin{align}
\tilde{\mathcal{H}}_{ijkl} &= \delta_{ik}\delta_{jl} \sum_{n=1}^N f''_{\hat{v}_{in}}(x_{in})x_{jn}^2 \\
&= 2\delta_{ik}\delta_{jl} \sum_{n=1}^N (\frac{1}{\hat{v}_{in}} + \frac{1}{x_{in}^2})x_{jt}^2.
 \label{eq:approx}
\end{align}
Our approximation provides an even sparser version of the true Hessian.
Then, then proposed update for the transform reads:
\begin{equation}
    \boxed{\mathbf{\Phi} \leftarrow \exp(-\eta \Pi_A(\tilde{\mathcal{H}}^{-1}\mathbf{G}))
    \mathbf{\Phi}}
    \label{eq:iterations}
\end{equation}
where $\eta$ is a step size.
The step size is chosen to verify the Wolfe conditions~\cite{wolfe1969convergence} and is computed using the classical interpolation algorithm thoroughly described in~\cite[pp.~59-60]{nocedal1999optim}.
%
Informally, Wolfe conditions guarantee that the objective function is sufficiently decreased by the step size, and that the projected gradient in the search direction is also decreased.
These conditions are critical to obtain convergence of quasi-Newton methods, and in practice help in achieving fast convergence.
Denote by $\tilde{\boldsymbol{\mathit{H}}}$ the matrix with coefficients $\tilde{\mathit{h}}_{ij} = 2 \sum_{n=1}^N (\frac{1}{\hat{v}_{in}} + \frac{1}{x_{in}^2})x_{jt}^2$, so that $\tilde{\mathcal{H}}_{ijkl} = \delta_{ik}\delta_{jl}\tilde{h}_{ij}$.
Our quasi-Newton's method solves all the aforementioned problems of Newton's method. The approximated Hessian is
\begin{itemize}
\item \textbf{cheap}: computing $\tilde{\mathcal{H}}$ has the same complexity as computing a gradient, i.e., $O(M^2\times N)$.
\item \textbf{positive definite:} the approximation boils down to a diagonal operator, i.e., $\tilde{\mathcal{H}}\mathbf{E} = {\tilde{\boldsymbol{\mathit{H}}}} \circ \mathbf{E}$.
Hence, its eigenvalues are the coefficients $\tilde{h}_{ij}$, which are all nonnegative.
As such, our method does not require Hessian regularization.

\item \textbf{easy to invert:} because it boils down to a diagonal operator, we have $\tilde{\mathcal{H}}^{-1}\mathbf{G} = \tilde{\boldsymbol{\mathit{H}}}^{\circ -1} \circ \mathbf{G}$.
Inversion is $O(M^2)$, which is negligible compared to the cost of computing the gradient.
\end{itemize}
The resulting optimization procedure is described in Algorithm~\ref{algo:optim}.

\begin{algorithm}[tb]
\SetKwInOut{Input}{Input}
\SetKwInOut{Output}{Output}
 \Input{Current factorization $\hat{\mathbf{V}}$, frames matrix $\mathbf{Y}$, current transform $\mathbf{\Phi}$, number of iterations $L$.}
 \For{l=1,\dots,L}{
	Compute $G$ and $\tilde{\mathcal{H}}$ using Eqs.~\eqref{eq:gradient}, ~\eqref{eq:approx} \\
	Compute the search direction $\mathbf{E} = -\Pi_A(\tilde{\mathcal{H}}^{-1} \mathbf{G})$ \\
	Compute a step size $\eta > 0$ satisfying the Wolfe conditions.\\
	Update $\mathbf{\Phi} \leftarrow \exp(\eta\mathbf{E}) \mathbf{\Phi}$
  }
 \Output{New transform $\mathbf{\Phi}$}
 \caption{Algorithm $\mathcal{A}$: Fast transform learning}
 \label{algo:optim}
\end{algorithm}

\subsection{Relation to independent component analysis (ICA)}

This objective function \eqref{eq:cost} is reminiscent of maximum-likelihood ICA where the maximum-likelihood objective is given by~\cite{pham1997blind}:
\begin{equation}
    \mathcal{L}(\mathbf{\Phi}) = - N \log |\det(\mathbf{\Phi})| + \sum_{m=1}^{M}\sum_{n=1}^N f([{\bf \Phi Y}]_{mn}),
\end{equation}
where $f$ is a pre-specified function.
Under the orthogonal constraint, $\log |\det(\mathbf{\Phi})| $ becomes constant. As such, the ICA objective function shares the same dependency in $\mathbf{\Phi}$ with TL-NMF and the algorithm proposed in this paper is inspired by the ICA acceleration techniques proposed in~\cite{ablin2018orthogonal}.
\section{Experiments}
\label{sec:expe}


%
The following experiments are run on a single core of a laptop equipped with an Intel Core i7-6600U @ 2.6 GHz processor and 16 GB of RAM. The Python code is available online.\footnote{\tt{https://github.com/pierreablin/tlnmf}}

\subsection{Synthetic data}
\label{sec:synth}
\begin{figure}
    \centering
    \includegraphics[width=0.9\columnwidth]{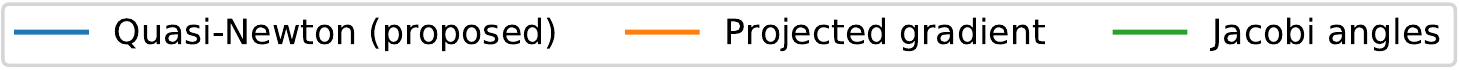} \\
    \begin{subfigure}[b]{0.32\columnwidth}
        \includegraphics[width=\columnwidth]{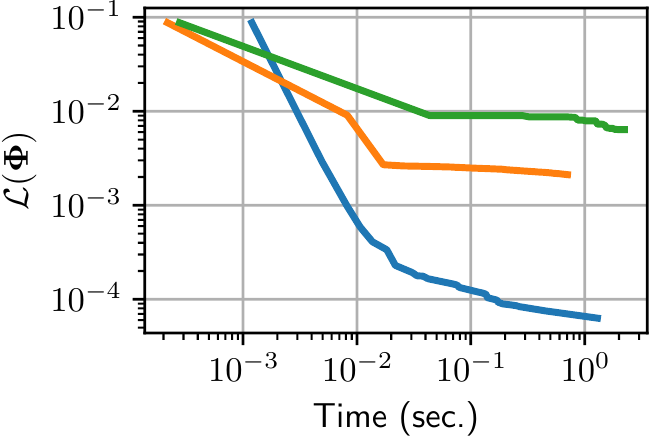}
        \caption{$M=10$}
    \end{subfigure}
    \begin{subfigure}[b]{0.32\columnwidth}
        \includegraphics[width=\columnwidth]{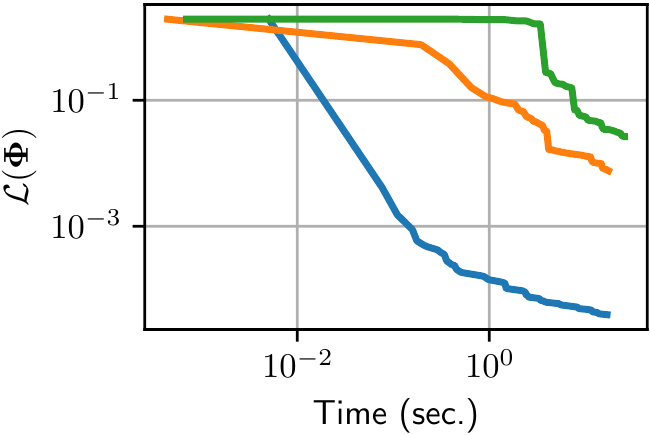}
        \caption{$M=100$}
    \end{subfigure}
        \begin{subfigure}[b]{0.32\columnwidth}
        \includegraphics[width=\columnwidth]{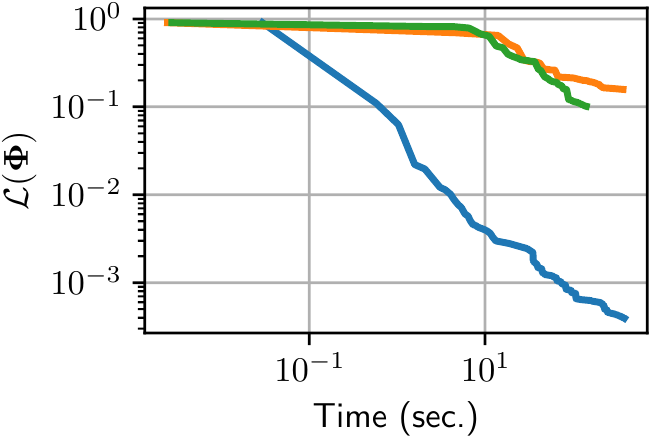}
        \caption{$M=500$}
    \end{subfigure}
    \caption{Convergence curves with synthetic data.}
    \label{fig:synth}
\end{figure}

We first focus on the sole optimization of $\mathcal{L}$, and not on the full TL-NMF procedure.
For this experiment, we generate random normal matrices $\bf Y$ of size $M \times N$, for $N=1000$ and $M\in [10, 100, 500]$, and a random transform $\mathbf{\Phi}^* \in \mathcal{O}_{M}$.
We set $\hat{\mathbf{V}} = |\mathbf{\Phi}^*\mathbf{Y}|^{\circ 2}$, so that the minimum of $\mathcal{L}(\bf \Phi)$ is $0$.
Algorithms start from an orthogonal initialization $\mathbf{\Phi}^0$ in the vicinity of $\mathbf{\Phi}^*$. More precisely, we set $\mathbf{\Phi}^0 =\exp(\mathbf{E})\mathbf{\Phi}^*$ where $\mathbf{E} = 10^{-3}\Pi_A(\mathcal{N}(0, \mathbf{I}_M))$.
Fig.~\ref{fig:synth} shows the convergence curve of the proposed method, projected gradient \cite{fagot2018transform} and Jacobi search \cite{wendt2018jacobi}. The proposed quasi-Newton approach leads to a drastic improvement in speed of convergence.

\subsection{Real data}
\label{sec:expe_tl}
{\noindent \bf Experimental setup.} We consider a $108$ seconds-long excerpt from \emph{My Heart (Will Always Lead Me Back To You)} by Louis Armstrong and His Hot Five. The sampling rate is $f_s = 11025$ Hz.
Using a $40$ ms-long analysis windows ($M=440$) with 50\% overlap between two frames, we obtain $N=5407$.
The rank of the decomposition is fixed to $K=10$, which is known empirically to provide a satisfactory decomposition with traditional NMF \cite{fevotte2009nonnegative}.\\

\begin{figure}
    \centering
    \includegraphics[width=0.9\columnwidth]{legend.pdf} \\
    \includegraphics[width=0.49\columnwidth]{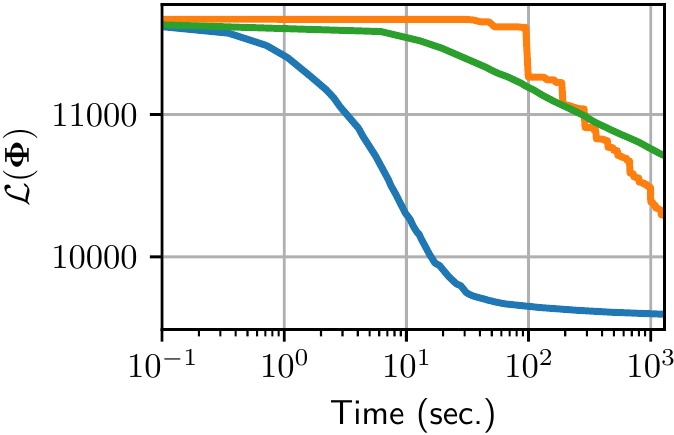}
    \includegraphics[width=0.45\columnwidth]{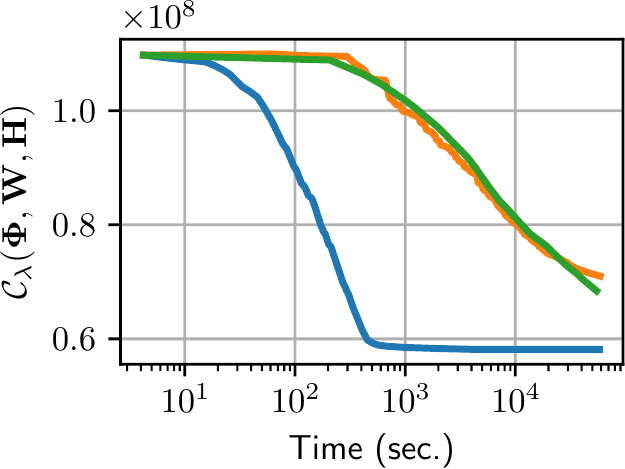}
    \caption{Convergence curves with real data. Left: minimization of $\mathcal{L}(\ve{\Phi})$ only. Right: full TL-NMF optimization.}
    \label{fig:single_problem}
\end{figure}

{\noindent \bf Comparison of the algorithms performance.}
In a first experiment, we first run traditional IS-NMF on the DCT spectrogram of the input signal and store $\hat{\mathbf{V}}$. Then the three transform learning algorithms are run with fixed $\hat{\mathbf{V}}$ and from a random starting point for $\ve{\Phi}$. This provides a realistic setting to compare their performance in optimizing  $\mathcal{L}(\mathbf{\Phi})$. Full TL-NMF (with free $\ve{W}$ and $\ve{H}$) are computed in a second experiment, using (same) random starting points. The three different transform learning algorithms are run with $L=5$. Results for the two experiments are shown in Fig.~\ref{fig:single_problem} and illustrate the superiority of the proposed quasi-Newton algorithm.\\

%
%
%

%

%

%
We now discuss some features of the transform learned with (full) TL-NMF using the quasi-Newton algorithm.
We will refer to the rows $\boldsymbol\phi_1, \cdots, \boldsymbol\phi_M$ of $\mathbf{\Phi}$ as \emph{atoms} (real-valued vectors of size $M$).
The learned atoms are not shown here due to space limitation but are similar to those obtained in~\cite{fagot2018transform, wendt2018jacobi}.\\

\begin{figure}
    \centering
    \begin{subfigure}[b]{0.55\columnwidth}
        \includegraphics[width=\columnwidth]{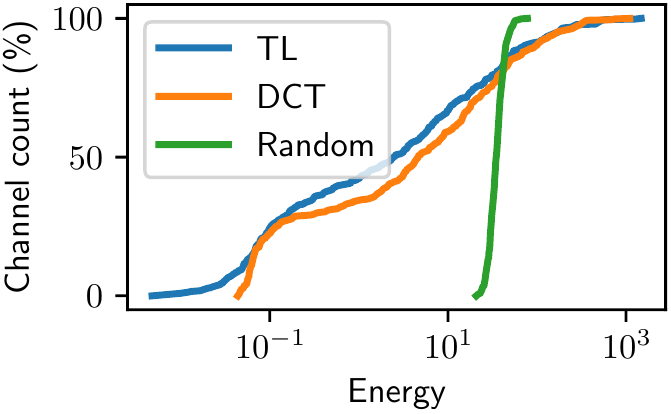}
        \caption{}
        \label{fig:energy}
    \end{subfigure}
    \begin{subfigure}[b]{0.44\columnwidth}
        \includegraphics[width=0.8\columnwidth]{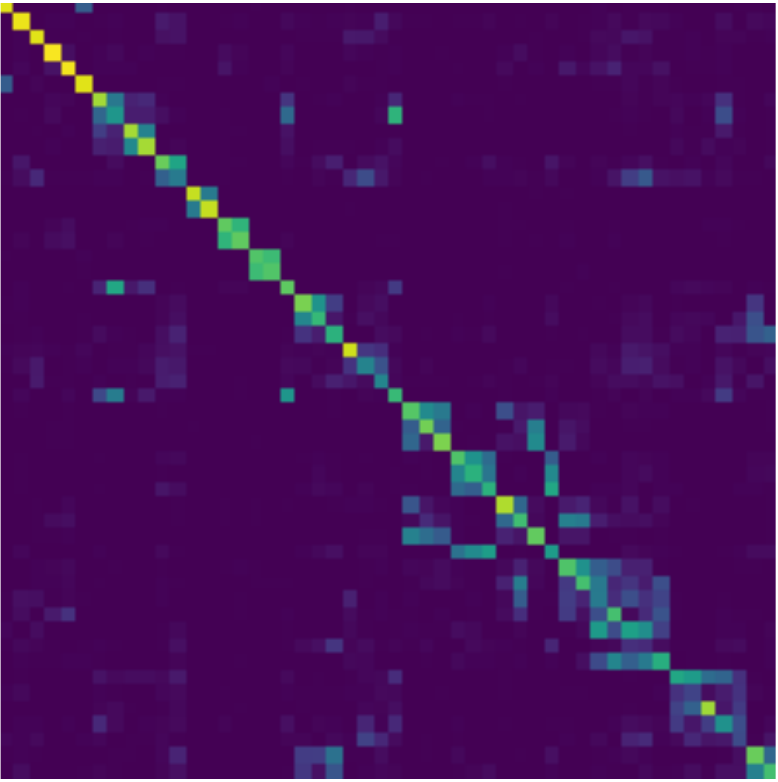}
        \includegraphics[width=0.122\columnwidth]{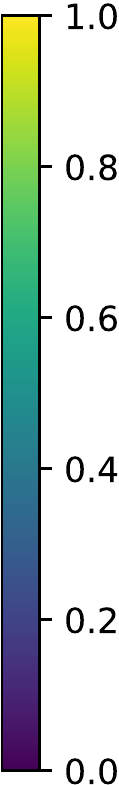}
        \caption{}
        \label{fig:init}
    \end{subfigure}
    \caption{\textbf{(a)}: Cumulative distribution of the atoms contributing energies for three transforms $\mathbf{\Phi}$. 
    \textbf{(b)}: Similarity matrix between the 50 most-contributing atoms learnt from two different random initializations. 
    }
    \label{fig:prop}
\end{figure}

{\noindent \bf Energy concentration.}
The contributed \emph{energy} of a single atom $\boldsymbol\phi_i$ is defined as $e_i = \sum_{n=1}^N [\boldsymbol\phi_i \mathbf{Y}]_n^2 $.
Fig.~\ref{fig:energy} shows the cumulative distribution of the energies for three different transforms: $\ve{\Phi}$ estimated by TL-NMF, the DCT, and a random orthogonal matrix.
As expected, DCT concentrates the energy while a random orthogonal transform barely does so. The energy concentration phenomenon is accentuated by transform learning.
This behavior was observed with other music datasets as well. \\
%

%
{\noindent \bf Reliability of the learned transform.} The problem solved by TL-NMF is non-convex, hence different initializations can lead to different local minima.
We investigate the structure of the local minima returned by the proposed quasi-Newton algorithm using a technique similar to ICASSO in ICA~\cite{himberg2004validating}.
It appears that a subset of atoms are reliably returned by the algorithm, regardless of initialization.
To observe this behavior, we consider two transforms obtained from two random initializations.
We select the 50 most-contributing atoms based of the values of $e_i$, yielding two matrices $\uline{\mathbf{\Phi}}^1$ and $\uline{\mathbf{\Phi}}^2$ of size $50 \times 440$.
We compute the correlation matrix $\mathbf{T} = \uline{\mathbf{\Phi}}^1 {\uline{\mathbf{\Phi}}^2}^{\top}$ of size $50 \times 50$ and find a permutation matrix $\mathbf{P}$ such that $\mathbf{P}\mathbf{T}$ is as block-diagonal as possible.
The absolute value of the resulting matrix is displayed in Fig.~\ref{fig:init}.
It is well structured and shows in particular that the first $6$ atoms (top left) are the same.
Furthermore, some pairwise couplings are also uncovered. The diagonal blocks in Fig.~\ref{fig:init} correspond to sets of atoms such that $\text{Span}(\boldsymbol\phi^1_i, \boldsymbol\phi^1_j) = \text{Span}(\boldsymbol\phi^2_{i'}, \boldsymbol\phi^2_{j'})$.

%
%
%
%

\section{Conclusion}

We introduced a quasi-Newton method on the orthogonal manifold to solve the TL-NMF problem.
It relies on a sparse approximation of the Hessian.
%
The proposed method outperforms the state-of-the-art methods by orders of magnitude.
%
On the laptop used for the experiments, the whole estimation took about 10 minutes for a $\sim$2-minutes signal, while NMF without transform learning takes roughly 2 minutes.
This work is thus a step towards making TL-NMF a practical tool for music signal processing.
The shortened time of estimation also helps investigate properties of the learned transform without prohibitive computational burden.
Results on the concentration of energy obtained by TL-NMF suggest that an algorithm that only learns a few atoms instead of $M$ would not result in too much loss of information.
Such an algorithm would further reduce the computational cost of TL-NMF, since the number of parameters would plummet.
We intend to study this matter in a future work.
\newpage

\bibliographystyle{IEEEbib}
\bibliography{refs}

\end{document}